\def\year{2018}\relax
\documentclass[letterpaper]{article} %DO NOT CHANGE THIS
\usepackage{aaai18}  %Required
\usepackage{times}  %Required
\usepackage{helvet}  %Required
\usepackage{courier}  %Required
\usepackage{url}  %Required
\usepackage{graphicx}  %Required

\usepackage{latexsym}
\usepackage{amsmath}
\usepackage{xcolor}
\usepackage{rotating}
\usepackage{subcaption}

\begin{document}
\title{Probabilistic Relation Induction in Vector Space Embeddings}
\author{Zied Bouraoui\\
Cardiff University, UK\\
BouraouiZ@Cardiff.ac.uk\\
\And
Shoaib Jameel\\
Cardiff University, UK \\
JameelS1@Cardiff.ac.uk\\
\And
Steven Schockaert\\
Cardiff University, UK\\
SchockaertS1@Cardiff.ac.uk
}

\maketitle
\begin{abstract}
Word embeddings have been found to capture a surprisingly rich amount of syntactic and semantic knowledge. However, it is not yet sufficiently well-understood how the relational knowledge that is implicitly encoded in word embeddings can be extracted in a reliable way. In this paper, we propose two probabilistic models to address this issue. The first model is based on the common relations-as-translations view, but is cast in a probabilistic setting. Our second model is based on the much weaker assumption that there is a linear relationship between the vector representations of related words. Compared to existing approaches, our models lead to more accurate predictions, and they are more explicit about what can and cannot be extracted from the word embedding. 
\end{abstract}

\section{Introduction}
A wide variety of methods have been proposed for representing words % (and related objects such as documents, phrases and entities) 
in low-dimensional vector spaces \cite{ASI:ASI1,Turney10,mikolov2013linguistic,glove2014}. While the primary motivation for most of these works has been to model similarity, recently there has been an increasing interest in the use of such vector space embeddings for learning other types of lexical relations. In particular, it has been observed that many syntactic and semantic relationships can be modelled as vector translations. For instance, there may be a vector $r$ that models the `captial of' relation, such that e.g.\ $p_{\textit{paris}}-p_{\textit{france}} \approx p_{\textit{tokyo}}-p_{\textit{japan}} \approx ... \approx r$, where $p_w$ denotes the representation of word $w$.

Although this remarkable property of word embeddings is now well-established, so far it has mainly been used as a tool for evaluating the quality of word embedding models \cite{glove2014,Vylomova2016}. In particular, it remains unclear to what extent, or in what way, vector space embeddings can be used for learning relations in tasks such as knowledge base completion \cite{NIPS20135071,west2014knowledge}.  To address this question, we focus on the following relation induction problem: given a set $\{(s_1,t_1),...,(s_n,t_n)\}$ of entity pairs that are related in a given way, %(e.g.\ city $t_i$ is the capital of country $s_i$), 
identify new entity pairs $(s,t)$ that are likely to be related in the same way. Throughout the paper, we will refer to $s$ and $t$ as the source and target word respectively. 

The context of knowledge base completion affects the relation induction task in a number of ways. First, it means that we need a model that can produce faithful confidence scores. This is important because adding incorrect information to a deductive system can have far-reaching consequences. Having faithful confidence scores means that we can repair any inconsistencies that arise in an informed way, and that we can qualify inference results that rely on automatically learned pieces of knowledge. Second, the number of training pairs $n$ is typically quite small, which means that common approaches, such as training a support vector machine (SVM) on the set of vector differences $t_i-s_i$ \cite{Vylomova2016}, may not be an ideal solution.

In this paper, we propose a probabilistic model that relies on two main ideas. First, it assumes that the sets of valid source and target words can be modeled as Gaussians. This prevents the model from identifying pairs $(s,t)$ in which $s$ or $t$ are not of the correct type (e.g.\ identifying the capital of something that is not a country), which addresses an important limitation of pure translation based models.  Second, it assumes that the set of translations $t-s$ which correspond to valid word pairs can also be modelled as a Gaussian. By considering a probability distribution over possible translations, rather than a single translation, the model is intuitively able to ignore features of word meaning that are irrelevant for the considered relation and to appropriately weight the remaining features. %This is similar in spirit to projecting the vectors on some relation-specific subspace, as has been advocated by several authors \cite{DBLP:conf/acl/RotheS16,roller2014inclusive}. 
We also consider a variant of our model that is not based on translations, and merely assumes that there is a linear mapping between source and target words, which we formalize using Bayesian linear regression. %Crucially, since both models produce faithful confidence scores, they can straightforwardly be used in combination, where a pair $(s,t)$ is predicted as soon as one of the models provides sufficient supporting evidence. %This variant is thus aimed to be complementary to the translation based model, focusing on types of semantic information that cannot be captured by translations.
% Explain intuitions of concatenation and regression model (wine-food pairing) but without talking about specific models (???)

%******************************************************************************************************************************
\section{Related Work}

\subsection{Predicting Relations}
At least three different types of approaches have been studied for predicting relations that are missing from a given knowledge base. First, there is a large body of work on relation extraction from text. %, going back at least to the seminal work of Hearst, which showed how hypernym relations could be effectively learned from a small set of simple linguistic patterns \cite{hearst1992automatic}. 
Among others, in recent years a number of approaches have been developed that are specifically targeted at completing knowledge bases. These methods essentially learn how to extract the considered relations by using their known instances as a form of distant supervision \cite{mintz2009distant,DBLP:conf/pkdd/RiedelYM10,surdeanu2012multi}. 

The second type of approaches rely on modeling statistical dependencies among the known instances of the considered relations, e.g.\ if we already know that ``person A works for company B'' and that ``company B is based in country C'', we can plausibly derive that ``A lives in C''. To exploit such dependencies, some approaches rely on learning latent representations \cite{Kok:2007:SPI:1273496.1273551,Speer:2008:ARD:1619995.1620084,Nickel:2012:FYS:2187836.2187874,DBLP:conf/naacl/RiedelYMM13,NIPS20135071,DBLP:conf/aaai/WangZFC14,yang2014embedding}, while others learn probabilistic rules \cite{DBLP:conf/emnlp/SchoenmackersDEW10,lao2011random,DBLP:journals/ml/WangMLC15}. 

The third type of approaches, which are the focus of this paper and are reviewed in more detail below, rely on vector space representations of words or entities to induce plausible relation instances. These vector space representations summarize the linguistic contexts in which the words/entities occur, and relations are thus essentially induced by comparing linguistic contexts. A standard approach is to model relations as translations in the vector space \cite{mikolov2013linguistic}, although various other approaches have also been investigated \cite{weeds2014learning}. 

These three types of methods are highly complementary. While relation extraction methods can predict very fine-grained relations, they require that at least one sentence in the corpus states the relation explicitly. Statistical methods can predict relations even without access to a text corpus, but they are limited to predicting what can plausibly derived from what is already known. %, and cannot e.g.\ predict relations involving entities that do not appear in the initial knowledge base. 
%Word embeddings can reveal relations that cannot be predicted using the other methods at all, such as syntactic relationships (e.g.\ relating the present and past tense of a verb). 
From a knowledge base completion point of view, the main appeal of word embeddings is that they may be able to reveal commonsense relationships which are rarely stated explicitly in text.

%---------------------------------------------------------
\subsection{Modeling Relations in a Vector Space}

As already mentioned in the introduction, various syntactic and semantic relations can be modeled as vector translations in a word embedding \cite{mikolov2013linguistic}. Among others, it has been shown that word embeddings can be used to complete analogy questions of the form \textit{a}:\textit{b}::\textit{c}:?, asking for a word that relates to $c$ in the same way that $b$ relates to $a$ (e.g.\ \textit{france}:\textit{wine}::\textit{germany}:?), by predicting the word $w$ that maximizes $\cos(p_b-p_a+p_c,p_w)$. 

Several types of interpretable features can be modeled as directions in word embeddings. %For example, it has been observed that better results are sometimes obtained by completing the analogy \textit{a}:\textit{b}::\textit{c}:? with the word $w$ maximizing $\cos(b-a,w-c)$ \cite{levy2014linguistic}.
For example, in \cite{DBLP:conf/acl/RotheS16}, it was shown that word embeddings can be decomposed in orthogonal subspaces that capture particular semantic properties, including a one-dimensional subspace (i.e.\ a direction) that encodes polarity. Along similar lines, in \cite{kim2013deriving} it was found that the direction defined by a word and its antonym (e.g. ``good'' and ``bad'') can be used to derive adjectival scales (e.g.\ bad $<$ okay $<$ good $<$ excellent). In \cite{gupta2015distributional}, it was shown that many types of numerical attributes can be predicted from word embeddings (e.g.\ GDP, fertility rate and CO$_2$ emissions of a country) using linear regression, again supporting the view that directions can model meaningful relations. Finally, in \cite{derracAIJ} an unsupervised method was proposed to decompose domain-specific vector spaces into interpretable directions. %, modeling the salient properties of the domain. 
For instance, in a space of movies, directions modeling terms such as ``scary'', ``romantic'' or ``hilarious'' were found. 

Several authors have focused on extracting hyperpnym relations from word embeddings. In \cite{baroni2012entailment}, To decide whether a word $h$ is a hypernym of $w$, in \cite{baroni2012entailment} it is proposed to use an SVM with a polynomial kernel, using the concatenation of $h$ and $w$ as feature vector. %Furthermore, it is reported that using vector differences (i.e.\ translations) instead of concatenations does not lead to competitive results. 
In \cite{roller2014inclusive} it was shown that vector differences can lead to good results with a linear SVM, provided that the vectors are normalized, and that the squared differences of each coordinate are added as additional features. Intuitively, this allows the SVM classifier to express that $h$ and $w$ need to be different in particular aspects (using the vector differences) but similar in other aspects (using the squared differences). 
Some authors have also proposed to identify hypernyms by using word embedding models that represent words as regions or densities \cite{Erk:2009:RWR:1596374.1596387,DBLP:journals/corr/VilnisM14,DBLP:conf/conll/JameelS17}.
%In \cite{weeds2014learning} vector differences were found to perform well for recognizing hypernyms, but other choices, such as vector addition, performed better for recognizing co-hyponyms.

Beyond hypernyms, most work has focused on completing analogies. The problem of relation induction, where we are given a set of correct instances instead of just one in the analogy task, was studied in \cite{Vylomova2016}, where a linear SVM trained on vector differences was used. While strong results were obtained for several relations in a controlled setting (e.g.\ predicting which among a given set of relations a word pair belongs to), many false positives were obtained when random word pairs were added as negative examples. To alleviate this issue, a number of heuristics were proposed to generate more informative negative examples. %, leading to substantial improvements. 
A variant of the relation induction problem was also studied in \cite{DBLP:conf/coling/DrozdGM16}, where the focus was on predicting the target word $t$ given a valid source word $s$ (as in analogy completion), given a set of training instances (as in relation induction). Two strong baselines were introduced in this paper, which will be discussed below as part of our experimental methodology.

%Among others, it was shown that strong baseline results can be obtained by predicting the word that is closest to $s + \frac{1}{n}\sum_{i=1}^n(t_i-s_i)$, i.e.\ the relation is modelled as a vector translation which is estimated as the average of the differences $t_i-s_i$ of the training instances. Another suggestion is to use a logistic regression classifier to predict whether a given word $w$ is a valid target word (i.e.\ whether it has the correct semantic type), and to multiply the output of that classifier with the cosing similarity of $w$ and $s$.

%******************************************************************************************************************************
\section{Modeling Relations}

In this section, we propose two models for relation induction. %, which rely on different assumptions about how the considered relation can be represented in the word embedding. 
%In both cases, 
We assume that we are given a set of pairs $\{(s_1,t_1),...,(s_n,t_n)\}$ as training data, and we need to determine whether a given pair of words $(s,t)$ are related in the same way. %For a word $w$, we write $p_w$ for its representation the word embedding.

\subsection{Translation Model}

\begin{figure}
\centering
\includegraphics[width=165pt]{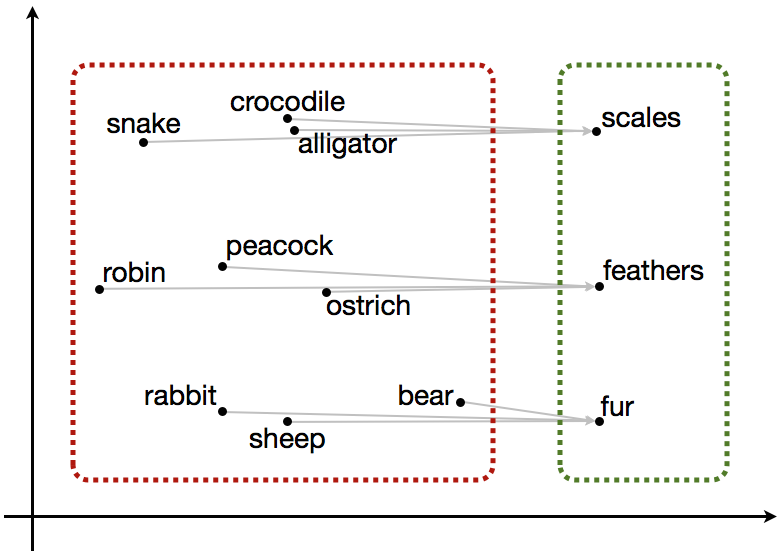}
\caption{Modeling relations as translations.\label{figTranslationModel}}
\end{figure}

The first model is based on the common view that relations can be modeled as vector translations. The source words $s_1,...,s_n$ typically belong to some semantic or syntactic category, and as a result their representations typically belong to some particular subspace of the word embedding. This is illustrated in Figure \ref{figTranslationModel} for the `has body covering' relation, where the source words all represent animals and the target words represent body covering types. 

If a relation can be modeled as a translation, it means that the source subspace and the target subspace have to be aligned. However, this is rarely perfectly the case. In fact, in most cases the source and target space even have a different number of dimensions. %, e.g.\ while a two-dimensional subspace may be enough to represent body coverings, a higher-dimensional space is needed to model the different semantic relations between animals. 
In the example from Figure \ref{figTranslationModel}, we can see that there is no vector that perfectly models the relation, although all valid word pairs $(s_i,t_i)$ define a translation which is more or less horizontal. This can be naturally modeled by representing the relation as a probability distribution over vector translations. In this example, this distribution would have a large `horizontal variance' but a very small `vertical variance'. Note that by considering probability distributions over translations, as special cases we can represent relations that are modeled as directions or in terms of similarity.

Intuitively, we want to accept $(s,t)$ as a valid relation instance if (i) the translation $t-s$ has a sufficiently high probability and (ii) $s$ and $t$ are of the correct type. Let us write $\delta_s$ and $\delta_t$ be the event that the source word $s$ and target word $t$ are of the correct type, and let $\theta_{st}$ be the event that $s$ and $t$ are in the considered relation. Note that $\theta_{st}$ entails $\delta_s$ and $\delta_t$.
We evaluate the probability that $(s,t)$ is a valid instance as follows:
\begin{align}
 &P(\theta_{st} | p_s,p_t) \notag\\ 
 &= P(\delta_s| p_s)  \cdot P(\delta_t| p_t) \cdot P(\theta_{st} | p_s,p_t,\delta_s,\delta_t)\notag\\
 &\propto \frac{f(p_s | \delta_s)}{f(p_s)} \cdot \frac{f(p_t | \delta_t)}{f(p_t)} \cdot \frac{f(p_t-p_s|\theta_{st})}{f(p_t-p_s | \delta_s,\delta_t)}\label{eqScoreTranslation}
\end{align}
To evaluate the latter expression we have to make a number of assumptions. First, we assume that the overall distribution of the words in the word embedding follows a multivariate Gaussian distribution. Given the typical vocabulary sizes, we can use the sample mean and covariance to estimate the parameters of this Gaussian, and thus evaluate $f(p_s)$ and $f(p_t)$. 
We also assume that $f(p_s | \delta_s)$ follows a multivariate Gaussian distribution. However, as the number of training instances $n$ is often small, and in particular smaller than the number of dimensions, the sample covariance matrix is not a reliable estimator. To alleviate this problem, we will restrict ourselves to diagonal covariance matrices. %However, since the estimation of $f(p_s | \delta_s)$ is then very sensitive to the direction of the axes, we first apply a coordinate transformation. Specifically, let $S$ be a matrix whose rows are the vector representations of the source words $s_1,...,s_n$ and let $S=U_S\Sigma_S V_S^T$ be the singular value decomposition (SVD) of $S$. Then we represent each source word using the vector $p_s^S = (p_s \cdot v_1^S,...,p_s \cdot v_m^S)$, where $v_1^S,...,v_m^S$ are the row vectors of $V_S$. 
We then have:
$$
f(p_s | \delta_s) = \prod_{i=1}^{m} f(x_i^s | \delta_s)
$$
where $m$ is the number of dimensions in the word embedding, 
%$x_i^s = p_s \cdot v_i^S$, 
$x_i^s$ is the $i^{\textit{th}}$ coordinate of $p_s$, 
and $f(x_i^s | \delta_s)$ follows a univariate Gaussian distribution with an unknown mean and variance. Using a Bayesian approach, we estimate $f(x_i^s |\delta_s)$ as:
$$
\int G(x_i^s;\mu,\sigma^2) \textit{NI$\chi^{2}$}(\mu,\sigma^2 | \mu_0,\kappa_0,\nu_0,\sigma_0^2) d\mu d\sigma
$$
where $G$ represents the Gaussian distribution and \textit{NI$\chi^{2}$} is the normal inverse $\chi^2$ distribution. This integral has an analytical solution, which is given as follows if we use flat priors on the parameters:
\begin{align*}
&t_{n-1}\left(\overline{x_i},\frac{(n+1)\sum_{j=1}^n (x_i^{s_j}-\overline{x_i})^2}{n(n-1)}\right)
\end{align*}
with $\overline{x_i} = \frac{1}{n}\sum_{j=1}^n x_i^{s_j}$ and $t_{n-1}$ the Student t-distribution with $n-1$ degrees of freedom.

The density $f(p_t | \delta_t)$ is evaluated in the same way. If we assume that the translations $p_t-p_s$ also follow a Gaussian distribution, we can estimate $f(p_t-p_s|\theta_{st})$ in a similar way. 
%In this case, we try to find a coordinate transformation that aligns the axes with the distribution of both the source and target words. Specifically, let $A$ be a matrix whose rows are the vector representations of the source and target words $s_1,..,s_n,t_1,...,t_n$, and let $A = U_{ST}\Sigma_{ST}V^T_{ST}$ be the singular value decomposition of $A$. Then we write $p_s^{ST}=(p_s \cdot v_1^{ST},...,p_s \cdot v_m^{ST})$ with  $v_1^{ST},...,v_m^{ST}$ the row vectors of $V_{ST}$. 
In particular, we estimate $f(p_t-p_s|\theta_{st})$ as $\prod_{i=1}^m f(x_i^s-x_i^t|\theta_{st})$, where $x_i^s$ is again the $i^{\textit{th}}$ coordinate of $p_s$ and similar for $x_i^t$. % = p_s \cdot v_i^{ST}$ and $y_i^t = p_t \cdot v_i^{ST}$. 
Each univariate Gaussian $f(x_i^s-x_i^t|\theta_{st})$ is then again estimated using the t-distribution, from the set of data points $\{x_i^{s_1}-x_i^{t_1},...,x_i^{s_n}-x_i^{t_n}\}$. We similarly estimate $f(p_t-p_s | \delta_s,\delta_t)$ as $\prod_{i=1}^m f(x_i^s-x_i^t|\delta_s,\delta_t)$. The mean of $f(x_i^s-x_i^t|\delta_s,\delta_t)$ is the same as the mean of $f(x_i^s-x_i^t|\theta_{st})$, but the variance is estimated from the differences $x_i^{s_l}-x_i^{t_k}$ corresponding to $n$ randomly sampled source words $s_l$ and target words $t_k$.

Note that if the assumption that the considered relation corresponds to a translation is wrong, we can expect the variance of $f(x_i^s-x_i^t|\theta_{st})$ and $f(x_i^s-x_i^t | \delta_s,\delta_t)$ to be similar, in which case the last factor in \eqref{eqScoreTranslation} evaluates to approximately 1. In other words, the model implicitly takes into account how much the translation assumption appears to be satisfied. %This allows us to easily different models, relying on different assumptions, without the risk of predicting too many false positives. 

\subsection{Regression Model}

\begin{figure}
\centering
\includegraphics[width=165pt]{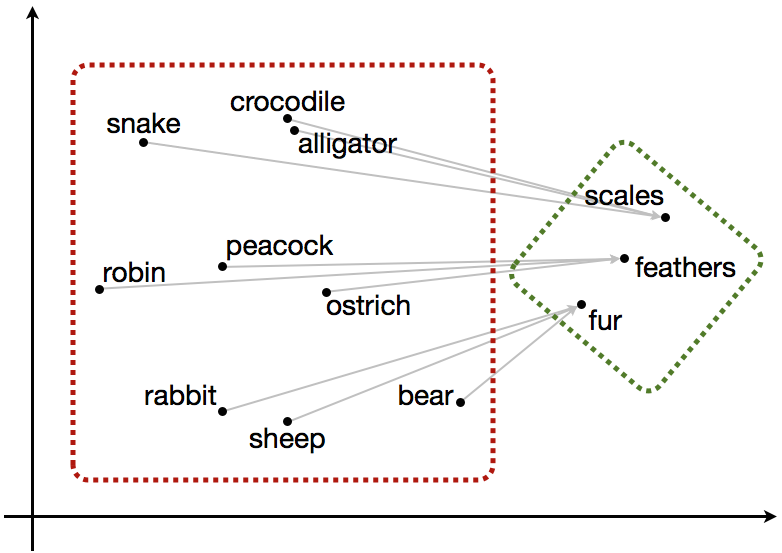}
\caption{Modeling relations as linear mappings.\label{figRegression}}
\end{figure}

The translation model relies on the assumption that the source and target spaces are aligned. For a relation such as `capital of', there is a direct connection between each source word and its corresponding target word, i.e.\ the representation of a country should be similar to the representation of its capital city. In such cases, we can indeed expect this alignment assumption to hold. There are many types of relations, however, for which the connection between source and target word is more implicit. Consider, for instance, the problem of wine-food pairing. We can expect that in a subspace with wines there are directions that correspond to features such as `sweetness', `acidity' and `amount of tannins'. Similarly, in a subspace of food types, there may be directions corresponding to features such as `healthy' or prototypes such as `meat', `fish' and `tomato'. The mere fact that such features are represented in the word embedding should be enough to predict reasonable wine-food pairings, even if the wine and food spaces are not aligned. 
Figure \ref{figRegression} illustrates this situation for the earlier example of animal body coverings. Clearly, the lack of alignment between the animal and covering spaces means that translation vectors are no longer a reliable indicator of whether the relation holds. Instead we have to rely on the weaker assumption that there is a linear mapping from the source to the target space. 

Taking this view, in this section we treat relation induction as a linear regression problem. However, two issues need to be addressed. First, we can only fit a linear regression model if the number of training examples is higher than the number of dimensions, which will often not be the case. We address this issue by using a low-rank approximation of the source space. Second, we need to explicitly represent how certain we are about the predictions of the linear regression model. If the source word $s$ is not a linear combination of the source words $s_1,...,s_n$ in the training data, then our model should capture the fact that the available training data is not sufficient to make a reliable prediction. Furthermore, even if $s$ is (approximately) a linear combination of $s_1,...,s_n$, we may only be able to predict particular features of the target space. To capture both sources of uncertainty, we make use of a Bayesian linear regression model.

In particular, we now model the probability that $(s,t)$ is a valid instance of the considered relation as follows: 
 \begin{align}
 &P(\theta_{st} | p_s,p_t) \notag\\ 
 &= P(\delta_s| p_s)  \cdot P(\delta_t| p_t) \cdot P(\theta_{st} | p_s,p_t,\delta_s,\delta_t)\notag\\
 &\propto \frac{f(p_s | \delta_s)}{f(p_s)} \cdot \frac{f(p_t | \delta_t)}{f(p_t)} \cdot \frac{f(p_t|p_s,\theta_{st})}{f(p_t |p_s, \delta_s,\delta_t)}\notag\\
 &= \frac{f(p_s | \delta_s)}{f(p_s)} \cdot \frac{f(p_t | \delta_t)}{f(p_t)} \cdot \frac{f(p_t|p_s,\theta_{st})}{f(p_t | \delta_t)} \notag\\
 &= \frac{f(p_s | \delta_s)}{f(p_s)} \cdot \frac{f(p_t|p_s,\theta_{st})}{f(p_t)} \label{eqScoreRegression}
\end{align}
 The densities $f(p_s | \delta_s)$, $f(p_s)$ and $f(p_t)$ are estimated as before. We estimate $f(p_t|p_s,\theta_{st})$ as $\prod_{i=1}^{m} f(x_i^t|p_s,\theta_{st})$, where $x_i^t$ is again the $i^{th}$ coordinate of $p_t$.
 %the vector representation of $t$, after doing the same SVD based coordinate transformation we use for estimating $f(p_t| \delta_t)$ (i.e.\ based on the decomposition of a matrix with $p_{t_1},...,p_{t_n}$ as row vectors). 
 
 Each univariate density $f(x_i^t|p_s,\theta_{st})$ is estimated using a Bayesian linear regression model that predicts the possible representations of the target word from $p_s$. % given a low-rank representation of the source word, as follows. 
 %The linear regression model for $f(z_i^t|p_s,\theta_{st})$ aims to predict $z_i^t$ from $p_s$. 
 However, this is only feasible if $p_s$ has at most $n-2$ coordinates. Therefore, we use a low-rank approximation of the source word representations, as follows.  Let $S$ be a matrix whose rows are the vectors $p_{s_1},...,p_{s_n}$ and let $A = U \Sigma V^T$ be the SVD decomposition of $A$. Let $v_1,...,v_k$ be the first $k$ row vectors of $V$, for some $k<n-1$. For a given vector $p$, we can think of $p^S=(p\cdot v_1,...,p\cdot v_k)$ as the representation of $p$ in the source subspace. Given that we typically need far fewer dimensions to represent the source space than the total number of dimensions in the word embedding, we should be able to predict the target word from $p_s^S$, even for relatively small values of $k$. In any case, the choice of $k$ represents a trade-off: the lower the value of $k$, the better we can characterize the uncertainty underlying our predictions, but the less information we have for making predictions. In the experiments, we have used $k=\frac{n-1}{2}$.
 %Let us write $p_{s_i}^s = (y_1^i,...,y_k^i)$. 
 We estimate $f(x_i^t|p_s,\theta_{st})$ as follows:  
\begin{align*}
\int &G(x_i^{t}; p_s^* \beta,\sigma^2) \cdot\\
 & G(\beta;(X^TX)^{-1}X^T b^i,(X^TX)^{-1}\sigma^2) \cdot\\
 & \textit{NI$\chi^{2}$}(\sigma^2 | \nu_0,\sigma_0^2) d\beta d\sigma
\end{align*}
where $b^i =(x_i^{t_1}, ..., x_i^{t_n})$, $X$ is composed of the first $k$ columns of $U\Sigma$ (with $U$ and $\Sigma$ the matrices from the SVD decomposition of $A$) with an additional 1 appended at the end of each row for the bias term, and $p_s^*$ is the vector $p_s^S$ with an additional 1 appended. Assuming a flat prior on the residual variance $\sigma^2$, the parameters $\nu_0$ and $\sigma_0^2$ can be estimated from the training data as:
\begin{align*}
\nu_0 &= n- k -1\\
\sigma_0^2 &= \frac{1}{n- k -1} (b^i - X\hat{\beta})^T(b^i - X\hat{\beta})
\end{align*}
with $\hat{\beta}$ the least squares solution.

\section{Evaluation}

In this section, we experimentally compare the two proposed models with a number of baseline methods from the literature. The relations we consider are taken from three standard benchmark datasets, each containing a mixture of syntactic and semantic relationships: (i) the Google Analogy Test Set (Google), which contains 14 types of relations with a varying number of instances per relation \cite{DBLP:journals/corr/abs-1301-3781}, (ii) the Bigger Analogy Test Set (BATS), wich contains 40 relations with 50 instances per relation \cite{DBLP:conf/naacl/GladkovaDM16}, and (iii) the DiffVec Test Set (DV), which contains 36 relations with a varying number of instances per relation \cite{Vylomova2016}.
%\begin{itemize}
% \item the Google Analogy Test Set (Google) \todo{add statistics} \cite{DBLP:journals/corr/abs-1301-3781};
%\item the Bigger Analogy Test Set (BATS) contains 40 relations with 50 instances per relation \cite{DBLP:conf/naacl/GladkovaDM16};
%\item the DiffVec Test Set (DV) contains 36 relations with a varying number of instances per relation \cite{Vylomova2016}.
%\end{itemize}
We report results for two embeddings that have been learned using Skip-gram, one from the  Wikipedia dump of  2 November 2015 (SG-Wiki) and one from a 100B words Google News data set\footnote{\url{https://code.google.com/archive/p/word2vec/}} (SG-GN). We also use two embeddings that have been learned with GloVe, one from the same Wikipedia dump (GloVe-Wiki) and one from the 840B words Common Crawl data set\footnote{\url{https://nlp.stanford.edu/projects/glove/}} (GloVe-CC).
%\begin{itemize}
%\item SG-Wiki: obtained using Skip-gram from the Wikipedia dump of November  2nd,  2015;
%\item GloVe-Wiki: obtained using GloVE from the Wikipedia dump of November  2nd,  2015;
%\item SG-GN: obtained using Skip-gram from a 100B words Google News data set\footnote{\url{https://code.google.com/archive/p/word2vec/}};
%\item GloVe-CC: obtained using GloVE from the 840B words Common Crawl data set\footnote{\url{https://nlp.stanford.edu/projects/glove/}}.
%\end{itemize}

For relations with at least 10 instances, we use 10-fold cross validation, whereas for relations with less than 10 instances, we use a leave-one-out evaluation. Note that the test fold only contains positive examples. To generate negative examples, we use four strategies. First, for each pair $(s,t)$ in the test fold, we add $(t,s)$ as a negative example. Second, for each source word $s$ in the test fold, we randomly sample two tail words from the test fold (provided that the test fold contains enough pairs), which do not occur together with $s$, and for each such tail word $t$, we add $(s,t)$ as a negative example. Third, for each positive example, we randomly select a pair from the other relations. Finally, for each positive example, we generate a random word pair from the words available in the dataset. This ensures that the evaluation involves negative examples that consist of related words, as well as negative examples that consist of unrelated words.

If we consider the task as a classification task, i.e.\ deciding for an unseen pair $(s,t)$ whether it has the considered relation, we need to select a threshold, as the considered methods only produce a confidence score (i.e.\ \eqref{eqScoreTranslation} for the translation model and \eqref{eqScoreRegression} for the regression model). To choose this threshold, we randomly select 10\% of the 9 training folds as validation data, and select the average score of the pairs just above and below the cut-off that optimizes the F1 score\footnote{Another possibility would be to choose priors that maximize the likelihood of the training data (and a random sample of negative examples). However, selecting a cut-off based on validation data allows for a more direct comparison with the baselines.}. In the results below, we separately report precision, recall and F1. We can also evaluate this task as a ranking problem, where we merely evaluate to what extent each method assigns the highest score to the correct pairs. In that case, we use mean average precision (MAP). %Note that the MAP score does not depend on the chosen threshold. 

%In addition to the translation and regression models, we also report results for a combined model, in which we take the maximum of the scores produced by our two models. Due to the nature of the models, this intuitively means that the translation based model will be used for relations where the underlying assumption is valid, while the regression model will be used otherwise.

\begin{table*}[t]
\centering
\footnotesize
\caption{Results of the relation induction experiments (macro-averages). \label{secResultsOverview}}
\begin{tabular}{|ll || @{\hspace{6pt}}c@{\hspace{6pt}}c@{\hspace{6pt}}c@{\hspace{6pt}} |  @{\hspace{6pt}}c@{\hspace{6pt}}c@{\hspace{6pt}}c@{\hspace{6pt}}   | @{\hspace{6pt}}c@{\hspace{6pt}}c@{\hspace{6pt}}c@{\hspace{6pt}} | @{\hspace{6pt}}c@{\hspace{6pt}}c@{\hspace{6pt}}c@{\hspace{6pt}} |}
\hline
&& \multicolumn{3}{c|@{\hspace{6pt}}}{SG-Wiki}& \multicolumn{3}{c|@{\hspace{6pt}}}{GloVe-Wiki}& \multicolumn{3}{c|@{\hspace{6pt}}}{SG-GN} & \multicolumn{3}{c|}{GloVe-CC}\\
&               & Google & BATS  & DV       & Google & BATS  & DV       & Google & BATS  & DV      & Google & BATS  & DV\\
\hline

\hline    
3CA & Pr        & 0.151 & 0.143 & 0.105        & 0.141 & 0.150 & 0.109        & 0.139 & 0.147 & 0.112            & 0.150 & 0.147 & 0.109    \\
3CA & Rec      & 0.737 & 0.776 & 0.585       & 0.724 & 0.776 & 0.547        & 0.725 & 0.777 & 0.600             & 0.732 & 0.763 & 0.582     \\
3CA & F1        & 0.251 & 0.242 & 0.178        & 0.235 & 0.251 & 0.182       & 0.234 & 0.247 & 0.189            & 0.249 &  0.246 & 0.183     \\
3CA & MAP     & 0.145 & 0.137 & 0.125      & 0.148 & 0.141 & 0.127        & 0.152 & 0.140 & 0.125               & 0.147 & 0.139 & 0.126     \\

\hline 
LRC & Pr        & 0.516 & 0.475 & 0.374          & 0.366 & 0.256 & 0.166       & 0.488 & 0.486 & 0.389              & 0.427 & 0.383 & 0.257   \\
LRC & Rec       & 0.646 & 0.672 & 0.527        & 0.527 & 0.577 & 0.439       & 0.670 & 0.646 & 0.570              & 0.659 & 0.596 & 0.474     \\
LRC & F1        & 0.573 & 0.557 & 0.437         & 0.432 & 0.355 & 0.241       & 0.565 & 0.555 & 0.462              & 0.518 & 0.466 & 0.333     \\
LRC & MAP       & 0.710 & 0.580 & 0.519       & 0.508 & 0.322 & 0.265       & 0.713 & 0.614 & 0.545              & 0.628 & 0.481 & 0.389     \\
\hline   
SVM & Pr        &  0.407  &   0.336 &  0.198     &  0.383 & 0.365 & 0.215       & 0.464 & 0.398  & 0.276             &   0.407 & 0.381  &  0.225    \\
SVM & Rec     & 0.680 & 0.417  &0.412         & 0.628 & 0.461 & 0.376        & 0.646 & 0.531 & 0.384              & 0.671 &  0.501  & 0.408     \\
SVM & F1        & 0.509 & 0.372  & 0.267          & 0.476 & 0.408 & 0.274        & 0.540 & 0.455  & 0.321             & 0.507 & 0.433  & 0.290     \\
SVM & MAP    & 0.494 & 0.366 & 0.283          & 0.502 & 0.404  & 0.298       & 0.611 & 0.467  & 0.366              & 0.502 & 0.425  & 0.296     \\
\hline
\hline     
Trans & Pr     & 0.794 & 0.627 & 0.449          & 0.635 & 0.445 & 0.284       & 0.741 & 0.660 & 0.498              & 0.744 & 0.571 & 0.378    \\
Trans & Rec    & 0.649 & 0.708 & 0.563           & 0.618 & 0.620 & 0.446       & 0.771 & 0.705 & 0.604              & 0.713 & 0.689 & 0.552     \\
Trans & F1     & 0.714 & 0.665 & 0.500           &  0.626 & 0.518 & 0.347       & 0.756 & 0.682 & 0.546              & 0.728 & 0.624 & 0.449     \\
Trans & MAP    & 0.906 & 0.729  & 0.596          & 0.791 & 0.541 & 0.387       & 0.890 & 0.773 &  0.635             &  0.898 & 0.678 & 0.520     \\
\hline 

Regr & Pr     & 0.668  & 0.474 & 0.410          & 0.536 & 0.281 & 0.259           &  0.627 & 0.476 & 0.469          & 0.613 &  0.401 & 0.357    \\
Regr & Rec    & 0.603 & 0.470 & 0.471          & 0.580 & 0.403 & 0.422           & 0.665 & 0.449 & 0.537           & 0.646 & 0.439 & 0.467   \\
Regr & F1     & 0.634 & 0.472  & 0.439          & 0.557 & 0.331 & 0.321           &  0.646 &  0.462 & 0.501       & 0.629 & 0.419 & 0.404    \\
Regr & MAP    &  0.834 & 0.618  & 0.570          & 0.741 & 0.434 & 0.381        &  0.801 & 0.639 &  0.621      & 0.793 &  0.549 & 0.506      \\
%\hline 
%Comb & Pr    & 0.794 & 0.626 & 0.454          & 0.330 & 0.233 & 0.165           & 0.740 & 0.659 &  0.508       & 0.744 & 0.576 & 0.384     \\
%Comb & Rec   & 0.649 & 0.706 & 0.601          & 0.539 & 0.582 & 0.500            & 0.771 & 0.712 &  0.616         & 0.713 & 0.686 & 0.542     \\
%Comb & F1    & 0.714 & 0.664 & 0.517          & 0.409 & 0.333 & 0.249            & 0.755 & 0.685  &  0.557        & 0.728 &0.626 & 0.449     \\
%Comb & MAP   & 0.843 & 0.732 & 0.564          & 0.475 & 0.283 & 0.229             & 0.836 & 0.752  &  0.590         & 0.846 & 0.668 & 0.483     \\
\hline
\end{tabular}
\end{table*}

\begin{sidewaystable*}[ph!]
\footnotesize
\centering
\caption{MAP scores for the individual relations of DiffVec and BATS for SG-GN}
\label{tabResultsBreakdown}
\begin{tabular}{|l | ccccc| }
\hline
DiffVec                                           & 3CA    & LRC   & SVM   & Trans & Regr  \\
\hline
Action:ObjectAttribute                   &    0.107     &    0.078    & 0.278      &  0.130  &  0.251      \\
Object:State                                  &   0.075     &   0.590    & 0.270      &   0.567  &  0.498     \\
Object:TypicalAction                     &   0.069     &   0.519    & 0.362      &  0.560  &   0.480     \\
Action/Activity:Goal                       &   0.122    &   0.490    & 0.290      &  0.515  &  0.475     \\
Agent:Goal                                    &   0.073     &   0.543    & 0.439      &  0.602  &   0.578     \\
Cause:CompensatoryAction         &   0.066     &   0.588    & 0.412      &  0.622  &  0.664     \\
Cause:Effect                                 &   0.103     &   0.464    & 0.241      &   0.484  &  0.444     \\
EnablingAgent:Object                  &   0.106     &   0.506    & 0.338      &   0.539  &  0.490     \\
Instrument:Goal                        &   0.094     &   0.371    & 0.232      &  0.419  &  0.437     \\
Instrument:IntendedAction           &   0.072     &   0.533    &  0.355      &  0.629  & 0.585     \\
Prevention                                   &   0.092     &   0.655    &  0.553     &  0.709  &  0.615     \\
Collective noun                            &   0.126     &    0.575    & 0.386      &  0.563  &  0.685     \\
Event                                        &   0.202     &   0.717   & 0.404      &  0.737  &   0.940     \\
Hyper                                       &   0.259     &   0.550    & 0.385      &  0.746  &  0.911     \\
Lvc                             		 &   0.072     &   0.709   & 0.772      &  0.735  &   0.220     \\
Mero                                            &   0.290     &   0.548    & 0.395      &  0.669  &  0.825     \\
Noun Singplur                              &   0.253     &    0.585    & 0.326      &  0.958  &   0.852     \\
Prefix re                                       &   0.204     &   0.497    & 0.307      &  0.721  &  0.689     \\
Concealment                               &   0.082     &   0.552    &  0.274      &   0.551  &  0.496     \\
Expression                                 &   0.055     &   0.818    & 0.508      &  0.810  &  0.822     \\
Knowledge                                  &   0.069     &   0.690    & 0.507     &  0.717  &  0.698     \\
Plan                                             &   0.083     &    0.548    & 0.285      &  0.566  &  0.622     \\
Representation                           &   0.100     &   0.500    & 0.398      &  0.485  &  0.388     \\
Sign:Significant                           &   0.093     &   0.384    &  0.301      &   0.393  &  0.380     \\
Attachment                                 &   0.091     &   0.542    & 0.234      &  0.654  &  0.526     \\
Contiguity                                   &   0.105     &   0.533    &  0.295      &  0.626  &  0.613     \\
Item:Location                              &   0.101     &   0.618    & 0.312      &  0.716  &  0.694     \\
Loc:Action/Activity                      &   0.076     &   0.736    &  0.511     &  0.757  &  0.727     \\
Loc:Instrument/AssociatedItem  &   0.075     &   0.476    & 0.334      &  0.487  &  0.500     \\
Loc:Process/Product                  &   0.107     &   0.407    & 0.569      &  0.478  &  0.656     \\
Sequence                                  &   0.111     &   0.405    &  0.271     &  0.424  &  0.409     \\
Time:Action/Activity                   &    0.101     &   0.559    & 0.347      &   0.554  &  0.606     \\
Verb 3rd                                    &   0.168     &   0.609    &  0.397      &  0.978  &  0.958     \\
Verb 3rd Past                           &   0.177     &   0.625    & 0.302      &  0.950  &  0.926     \\
Verb Past                                 &   0.185     &   0.638   & 0.316     &  0.985  &  0.901     \\
Vn-Deriv                                  &   0.349     &   0.445    & 0.259      &  0.828  &  0.792     \\
                                &        &       &       &       &          \\
                                &        &       &       &       &           \\
                                &        &       &       &       &           \\
                                &        &       &       &       &           \\
\hline
\end{tabular}
\begin{tabular}{|l | ccccc| }
\hline
BATS                               & 3CA    & LRC   & SVM   & Trans & Regr   \\
\hline
Regular plurals                     &   0.186     &   0.586    & 0.402      &  0.876  &  0.791     \\
plurals - orthographic changes      &   0.195     &   0.601    & 0.355     &  0.772  &  0.652     \\
Comparative degree                  &   0.106     &   0.654    & 0.473      &  0.961  &  0.883     \\
Superlative degree                  &   0.093     &   0.705    & 0.605      &  0.932  &  0.868     \\
Infinitive: 3Ps.Sg                  &   0.116     &   0.638    & 0.536      &  1.000  &  0.968     \\
Infinitive: participle              &   0.161     &   0.603    & 0.495     &  0.916  &  0.793     \\
Infinitive: past                    &   0.146     &   0.582    & 0.455      &  0.957  &  0.722     \\
Participle: 3Ps.Sg                  &   0.099     &   0.596    & 0.577      &  0.874  &  0.760     \\
Participle: past                    &   0.161     &   0.598    &  0.450      &  0.872  &  0.731     \\
3Ps.Sg: past                        &  0.131     &   0.696    &  0.556      &  0.984  &  0.949     \\

Noun+less                           &   0.079     &   0.582    &  0.433     &  0.620 &  0.634     \\
Un+adj                              &   0.110     &   0.556    & 0.354     &  0.774  &  0.692     \\
Adj+ly                              &   0.096     &   0.625    &  0.495     &   0.895 &  0.824     \\
Over+adh./Ved                       &   0.092     &   0.627    & 0.375      &   0.744  &  0.765     \\
Adj+ness                            &   0.077     &   0.717    & 0.568      &   0.838  &  0.832     \\
Re+verb                             &   0.138     &   0.662    & 0.376     &  0.828  &  0.746     \\
Verb+able                           &   0.085     &   0.628    & 0.569      &  0.758  &  0.762     \\
Verb+er                             &   0.075     &   0.646    &  0.613      &  0.793  &  0.700     \\
Verb+ation                          &   0.115     &   0.580    & 0.441      &  0.784  &  0.762     \\
Verb+ment                           &   0.099     &   0.543    &  0.474      & 0.782  &  0.694    \\

Hypernyms animals                 &   0.223     &   0.705    & 0.644     &  0.852  &  0.284     \\
Hypernyms misc                      &   0.202     &   0.649    & 0.545      &  0.778  &  0.239     \\
Hyponyms misc                       &   0.313     &   0.449    & 0.304     &  0.686  &  0.281     \\
Meronyms substance                  &   0.170     &   0.513    & 0.363      &  0.612  &  0.267     \\
Meronyms member                     &   0.134     &   0.558    & 0.325      &  0.682 &  0.660     \\
Meronyms part-whole                 &   0.255     &   0.525    & 0.391      &  0.650  &  0.259     \\
Synonyms intensity                  &   0.258     &   0.501    & 0.291      &  0.668  &  0.228     \\
Synonyms exact                      &   0.270     &   0.455    & 0.260      &  0.518  & 0.191     \\
Antonyms gradable                   &   0.251     &   0.487    & 0.316      &  0.567  &  0.276     \\
Antonyms binary                    &   0.219     &   0.440    &  0.306      & 0.497  &  0.239     \\

Capitals                            &   0.106     &   0.682    & 0.465      &  0.752  &  0.735     \\
Country:language                    &   0.088     &   0.643    & 0.524      &  0.668  &  0.717     \\
UK city: county                     &   0.053     &   0.785    & 0.587      &  0.877  &  0.719     \\
Nationalities                       &   0.059     &   0.774    & 0.603      &  0.852  &  0.626     \\
Occupation                          &   0.073     &   0.660    &  0.533      & 0.681  &  0.780     \\
Animals young                       &   0.115     &   0.601    & 0.458      &  0.687  &  0.754     \\
Animals sounds                      &   0.087     &   0.650    & 0.454     &  0.661  &  0.768     \\
Animals shelter                     &   0.116     &   0.665    & 0.558      &  0.679  &  0.501     \\
thing:color                         &   0.098     &   0.772    & 0.663      &  0.761  &  0.799     \\
male:female                         &   0.147     &   0.611    & 0.471      &  0.840  &  0.720     \\
\hline
\end{tabular}
\end{sidewaystable*}

\subsection{Baselines}\label{secBaselines}
The first baseline we consider is the 3CosAvg method proposed in \cite{DBLP:conf/coling/DrozdGM16}, which essentially treats the relation induction problem like an analogy completion problem, where we use the average translation vector across all pairs $(s_i,t_i)$ from the training data. In particular, this method assigns the following score to the test pair $(s,t)$:
\begin{align*}
\textit{score}_{\textit{3CA}}(t,s) = \cos\left(p_t,p_s + \frac{\sum_i p_{t_i} - p_{s_i}}{n}\right)
\end{align*}
Despite its simplicity, 3CosAvg was found to be a remarkably strong baseline. Another method proposed in \cite{DBLP:conf/coling/DrozdGM16}, called LRCos, is based on the assumption that $(s,t)$ is likely correct if $\cos(p_s,p_t)$ is high and $t$ is of the correct type, where a logistic regression classifier was trained on the target words $\{t_1,...,t_n\}$ to predict the probability that $t$ is a valid `target word'. To adapt this method to our setting, we also need to consider the probability that $s$ is a valid `source word' (which is not needed in the analogy completion setting considered in \cite{DBLP:conf/coling/DrozdGM16}, since there $s$ is always given as a valid source word). To allow for a more direct comparison with our methods, instead of using a logistic regression classifier, we will use our Bayesian estimation for the probability that $s$ and $t$ are of the correct type. In particular, we use the score $\textit{score}_{\textit{LRC}}(t,s)$ defined as follows:
\begin{align*}
\frac{P(p_s | \delta_s)}{P(p_s)} \cdot \frac{P(p_t | \delta_t)}{P(p_t)} \cdot \cos(p_s,p_t)
\end{align*}
As our final baseline, we train a linear SVM classifier using the training pairs $(s_1,t_1),...,(s_n,t_n)$ as positive examples. Following \cite{Vylomova2016}, we use negative examples of the form $(t_i,s_i)$, obtained by swapping the position of source and target word, as well as negative examples of the form $(s_i,t_j)$, obtained by swapping $t_i$ by the target word of another instance (while ensuring that $(s_i,t_j)$ does not appear in the training data as well). Finally, we also add $n$ random word pairs as negative examples. 
% \todo{Give details on how many negative examples of each type we add.} 
The $C$ parameter of the SVM is tuned for each relation separately (choosing values from $\{0.01,0.1,1,10,100\}$), by using the same validation data that is used for selecting the thresholds in the other models. To address class imbalance, negative examples were weighted by the ratio of positive to negative examples.

\subsection{Results}

The results are summarized in Table \ref{secResultsOverview}. As can be observed, our translation model consistently outperforms all other methods in both MAP and F1 score. Moreover, the regression model consistently outperforms the baselines in terms of MAP score, and outperforms the baselines in for the Google and DV test sets in terms of F1 score (but not for the BATS test set). Among the baselines, 3CosAvg is clearly outperformed by LRCos and SVM. On average, LRCos is the strongest baseline (except for the GloVe-Wiki embedding, where it is outperformed by SVM). This highlights the importance of explicitly modeling the fact that source and target words are expected to belong to a given type. Models which are only trained on translation vectors, such as in the SVM approach, cannot capture this. On the other hand, LRCos relies on cosine similarity to connect source and target words, which is far from optimal, as is evidenced by the large difference in performance between LRCos and our translation model.

\begin{figure*}[t]
\centering
\begin{subfigure}{.5\textwidth}
  \centering
  \includegraphics[width=.7\linewidth]{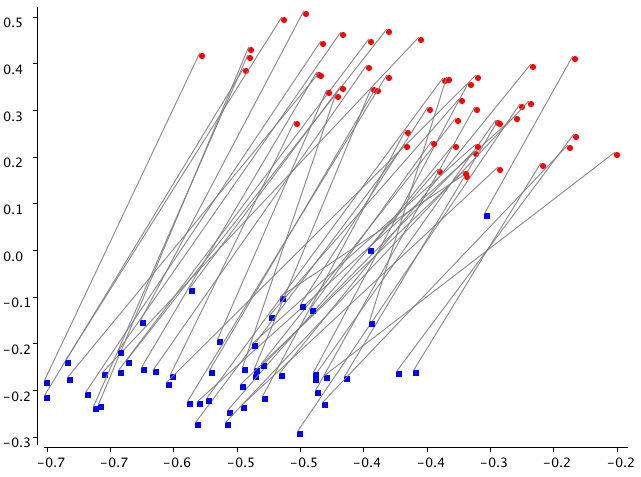}
  \caption{\label{figSuperlativeA}}
  \label{fig:sub1}
\end{subfigure}%
\begin{subfigure}{.5\textwidth}
  \centering
  \includegraphics[width=.7\linewidth]{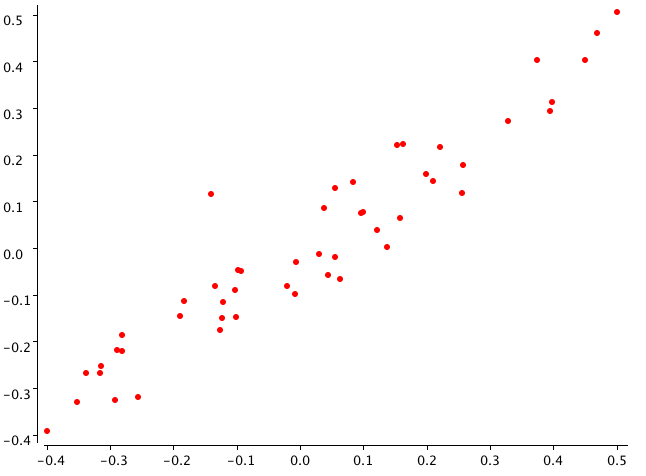}
  \caption{\label{figSuperlativeB}}
  \label{fig:sub2}
\end{subfigure}
\caption{Superlative degree relation of BATS for SG-GN. \label{figSuperlative}}
\label{fig:test}
\end{figure*}
 
\begin{figure*}[t]
\centering
\begin{subfigure}{.5\textwidth}
  \centering
  \includegraphics[width=.9\linewidth]{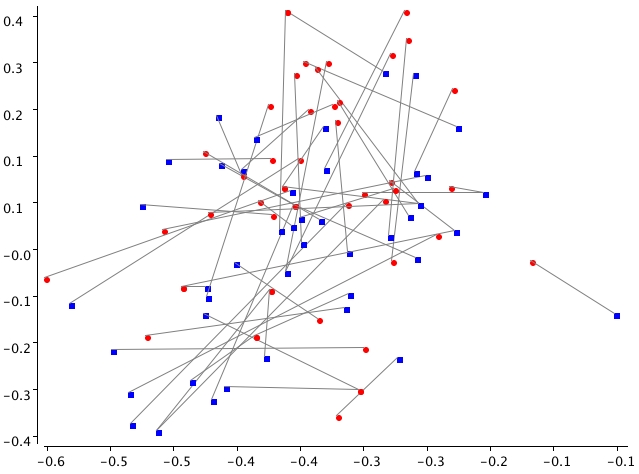}
  \caption{\label{figMeroA}}
  \label{fig:sub1}
\end{subfigure}%
\begin{subfigure}{.5\textwidth}
  \centering
  \includegraphics[width=.9\linewidth]{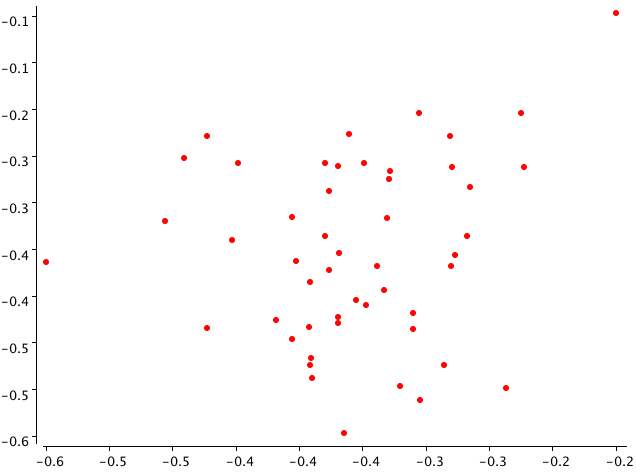}
  \caption{\label{figMeroB}}
  \label{fig:sub2}
\end{subfigure}%
\caption{Mero relation of DiffVec for SG-GN. \label{figMero}}
\label{fig:test}
\end{figure*}

To compare the performance of the methods across different types of relations,
% Table \ref{tabResultsBreakdown} contains the MAP scores for a number of selected relations from the DiffVec and BATS test sets, 
Table \ref{tabResultsBreakdown} contains the MAP scores for the relations from the DiffVec and BATS test sets, for the SG-GN word embedding. For the BATS dataset, the translation model consistently outperforms the baseline across all relations (including the relations that are not shown in the table). In the case of DiffVec there are a few exceptions, as can be seen in Table \ref{tabResultsBreakdown}, but in such cases the differences with the translation model are small. The regression model also outperforms the baselines in most cases, but there are a few exceptions where it performs much worse (e.g.\ \textit{Lvc} for DiffVec, and \textit{Hypernyms-animals}, \textit{Meronyms-substance}, \textit{Synonyms-intensity} and \textit{Antonyms-binary} in the case of BATS).

While the regression model is outperformed by the translation model on average, there are several cases where it performs better. For relations such as \textit{Event}, \textit{Hyper} and \textit{Mero} from DiffVec, where the number of examples is rather large (resp.\ 3583, 1173, 2825), we can see that the regression model actually substantially outperforms the translation model. The main weakness of the regression model is that it needs more training data: while a vector translation can be estimated from a single training example, learning an arbitrary linear mapping requires the number of training examples to be larger than the number of dimensions. While this can be addressed by using a low-dimensional approximation of the source word, doing so means that information is lost.

On the other hand, the main weakness of the translation model is that its underlying assumption is rather strong, and there are indeed some relations in the test sets that simply cannot be faithfully modeled in terms of translations. To illustrate this point, in Figure \ref{figSuperlative} we first show an example of a relation for which the translation assumption is clearly satisfied, the superlative relationship from the BATS test set. In particular, Figure \ref{figSuperlativeA} shows the first two principal components\footnote{Specifically, we obtained these coordinates based on an SVD decomposition of the representations of the relevant source and target words $\{s_1,...,s_n,t_1,...,t_n\}$.} of the representations of some word pairs $(s_i,t_i)$ that have this relationship (where related words are connected with a line). Clearly, this is the kind of plot that we would expect for a relation that satisfies the translation assumption. Figure \ref{figSuperlativeB} illustrates the same relation in a different way. Here a word pair $(s,t)$ is represented as a point $(x,y)$, where $x$ corresponds to the first principal component of the source word $s$, and $y$ corresponds to the first principal component of $t$. If the considered relationship satisfies the translation assumption, we would expect these points to lie on a line with a slope of 1, which is here (approximately) the case (as it is for other principal components).

Figure \ref{figMero} shows a similar plot to Figure \ref{figSuperlative} but for the \textit{mero} relation from the DiffVec test set. As is clearly illustrated by this figure, the translation assumption is not valid for this relation. The fact that the regression model performs quite well for this relation means that it can nonetheless be described using a linear model.

%The regression model again consistently outperform the other methods. However, for some relation, such as the lexicographic semantics relations of BATS for instance, the translation model perform better than the regression model. 

% \todo{add discussion about Table \ref{tabResultsBreakdown}}\\

% Explain that we evaluate as ranking and as classification
% Explain that both are in some sense more difficult than analogy completion where we simply have to find the most plausible target for each source word

% We call our model FAIRIE: Faithful Relation Induction from Embeddings

% Consider separate baselines for hypernyms? E.g. from \cite{roller2014inclusive}

% Difficulty with SVM : only synthetic negative examples

%******************************************************************************************************************************
\section{Conclusions}

We have proposed two probabilistic models for identifying word pairs that are in a given relation. The first model is based on the common assumption that lexical relations correspond to vector translations in a word embedding. The other model is based on linear regression, relying on the weaker assumption that there is a linear relationship between the source and target words of the considered relation. Both models implicitly factor in whether their underlying assumption is satisfied, and could thus easily be used in combination with each other, or with additional models. In our experimental evaluation, we have found both models to outperform existing approaches, with the translation model outperforming the regression model on average.

There are several interesting avenues for future work. First, a number of variants of the proposed models can be developed. For example, a model based on vector concatenations could intuitively model similar kinds of relationships as the regression model. However, in the case of vector concatenations, we can no longer use a diagonal covariance matrix, as that would mean that no interactions between source and target words are being captured. One solution could be to use a low-rank approximation of the vector concatenations and estimate full covariance matrices in a lower-dimensional space. Another interesting option to explore would be to estimate prior probabilities from coarser grained relations for which more training data is available. For example, we could learn a generic model for causal relations, and use that as a prior for the specific types of causal relationships that are considered in the DiffVec test set. It may even be useful to learn priors capturing e.g.\ syntactic relations, which would intuitively amount to finding a subspace of the embedding that relates to syntactic features.

%In addition to word embeddings, we will also consider a vector space embedding of WikiData entities \cite{DBLP:conf/ecai/JameelS16}, which will allow us to consider a wider variety of semantic relations (e.g.\ film/directed-by/person).

%\section*{Acknowledgments}

\bibliography{commonsense,wordembedding}
%\bibliography{entity,commonsense,wordembedding}
\bibliographystyle{aaai}

\end{document}